\documentclass[accepted]{uai2026} 
                        
\usepackage[american]{babel}

\usepackage{natbib} 
\usepackage{mathtools} 
\usepackage{booktabs} 
\usepackage{tikz} 

\usepackage{amsmath}
\usepackage{amssymb}
\usepackage{mathtools}
\usepackage{amsthm}
\usepackage{multirow}

\usepackage[capitalize,noabbrev]{cleveref}

\title{From Global to Factor-Wise Expert Composition in Discrete Diffusion Models}

\author[1,2]{\href{mailto:<marko@cs.toronto.edu>}{Haozhe~Huang}{}}
\author[2,3]{Yudong~W.~Xu}
\author[1,2]{Abhijoy~Mandal}
\author[1,2,4]{Al\'an~Aspuru-Guzik}

\newcommand{\acknowSciNet}[1][Trillium supercomputer]{Computations were performed on the #1 at the SciNet HPC Consortium.  SciNet is funded by Innovation, Science and Economic Development Canada; the Digital Research Alliance of Canada; the Ontario Research Fund: Research Excellence; and the University of Toronto.}

\affil[1]{%
    Department of Computer Science, University of Toronto
}
\affil[2]{
    Vector Institute for Artificial Intelligence
}
\affil[3]{
    Department of Mechanical \& Industrial Engineering, University of Toronto
}
\affil[4]{
    Senior Fellow, Canadian Institute for Advanced Research (CIFAR)
}
\begin{document}
\maketitle

\begin{abstract}
Discrete diffusion models offer a powerful framework for solving complex reasoning tasks, particularly through compositional generation, which combines multiple pre-trained experts to generalize beyond their individual training data. Recent theoretical corrections introduce time-dependent mixing weights to better align composed diffusion dynamics with the intended target. However, these methods are fundamentally limited by working on a per-sample basis, treating each generated state monolithically and ignoring the potential spatial or functional specializations of different experts. In this work, we address this limitation by proposing \textbf{FactorDiff} -- a factor-wise composition framework for diffusion models. We posit that samples can be further decomposed into smaller factors, and propose a sampling process that dynamically routes each factor to the most relevant expert. We instantiate this framework with spatial/pixel-level compositions and validate it on the ARC-AGI benchmark, demonstrating that simple factor-specific routing consistently outperforms complex global scalar weighting schemes on tasks that require logical consistency and spatial disentanglement.
\end{abstract}


\section{Introduction}
\label{sec:introduction}

Discrete diffusion models have recently emerged as a powerful paradigm for solving complex reasoning and generative tasks by learning the underlying data distribution directly from samples~\citep{d3pm2021,sedd2024,ye2025beyond}. A key advantage of this framework is its potential for compositional generation — the ability to combine multiple pre-trained models (or experts) to solve tasks that lie outside the training distribution of any single model~\citep{du2020compositional}. This capability is critical for real-world applications where collecting data for every possible combination of desired properties is intractable.


Recent works have sought to improve compositional sampling by introducing time-dependent correction weights for each expert. Methods such as SuperDiff~\citep{superdiff2024}, RNE~\citep{rne2025}, and Feynman-Kac Correctors~\citep{skreta2025feynmankac,feynmankac2025} derive these weights mathematically to ensure the combined score field approximates the true product distribution. While theoretically grounded, these approaches share a fundamental limitation: they assign a single global scalar weight to each expert at every timestep. This implicitly assumes that each expert is equally knowledgeable across the entire set of variables in the state space $x$.

This "monolithic" assumption is problematic for complex generative tasks where experts have specialized domains or conflicting objectives. For instance, in a spatial reasoning task, one expert may be valid only for a specific region, while another governs the boundary conditions. Assigning a single weight forces a compromise that dilutes the signal of the correct expert in its region of competence.

In this work, we argue that effective composition requires a more granular approach. We propose a factor-wise composition framework in which the state is decomposed into user-defined factors, and expert contributions are routed at the factor level rather than applied uniformly to the entire sample. Instead of assigning a single scalar weight for each expert, our method dynamically assigns weights per factor during sampling, enabling heterogeneous expert contributions within a single generated state. We instantiate this framework with position-level routing for grid-structured reasoning and demonstrate the benefits of this intuition on ARC-AGI, a challenging discrete-reasoning benchmark where standard scalar weighting fails to capture the necessary local constraints. By acknowledging the spatial and functional independence of experts, our method achieves superior generalization on tasks that require strict logical consistency.

\begin{figure*}[t]
    \centering
    \includegraphics[width=\textwidth]{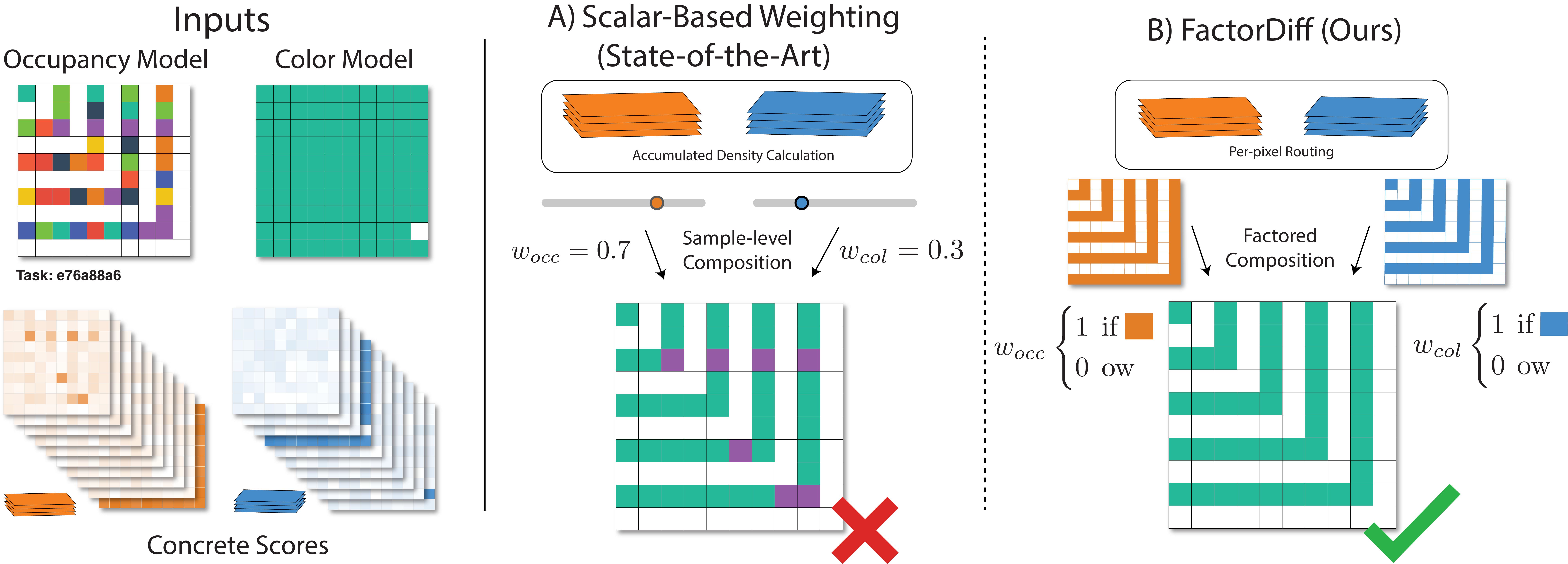}
    \caption{Overview of our method. Discrete diffusion models can be combined via their concrete scores during the reverse diffusion step. However, even state-of-the-art methods do not leverage them to their full potential, operating only at the sample level with scalar weights. By further factoring the sample space at the spatial level, we can produce better results with expert model combinations.}
    \label{fig:1}
\end{figure*}
We summarize our contributions as follows:
\begin{itemize}

    \item We propose \textbf{FactorDiff},\footnote{Code: \url{github.com/markohuang/factordiff}.} a factor-wise composition framework that casts composition as routing over factors, and instantiates it with position-level expert selection for grid-structured diffusion models (\cref{fig:1}).
    \item We introduce a native 2D masked diffusion architecture that preserves grid topology and enables training complementary expert models whose predictions can be composed at inference time.
    \item We empirically evaluate factor-wise routing on ARC-AGI, showing consistent gains over per-sample scalar-weight baselines (including SuperDiff, RNE, and Feynman--Kac correctors), with the largest improvements arising from highly specialized complementary experts.

    \end{itemize}

\section{Preliminaries}
\label{sec:background}

\subsection{ARC-AGI}
\label{sec:arc}

\begin{figure}
    \centering
    \includegraphics[width=\linewidth]{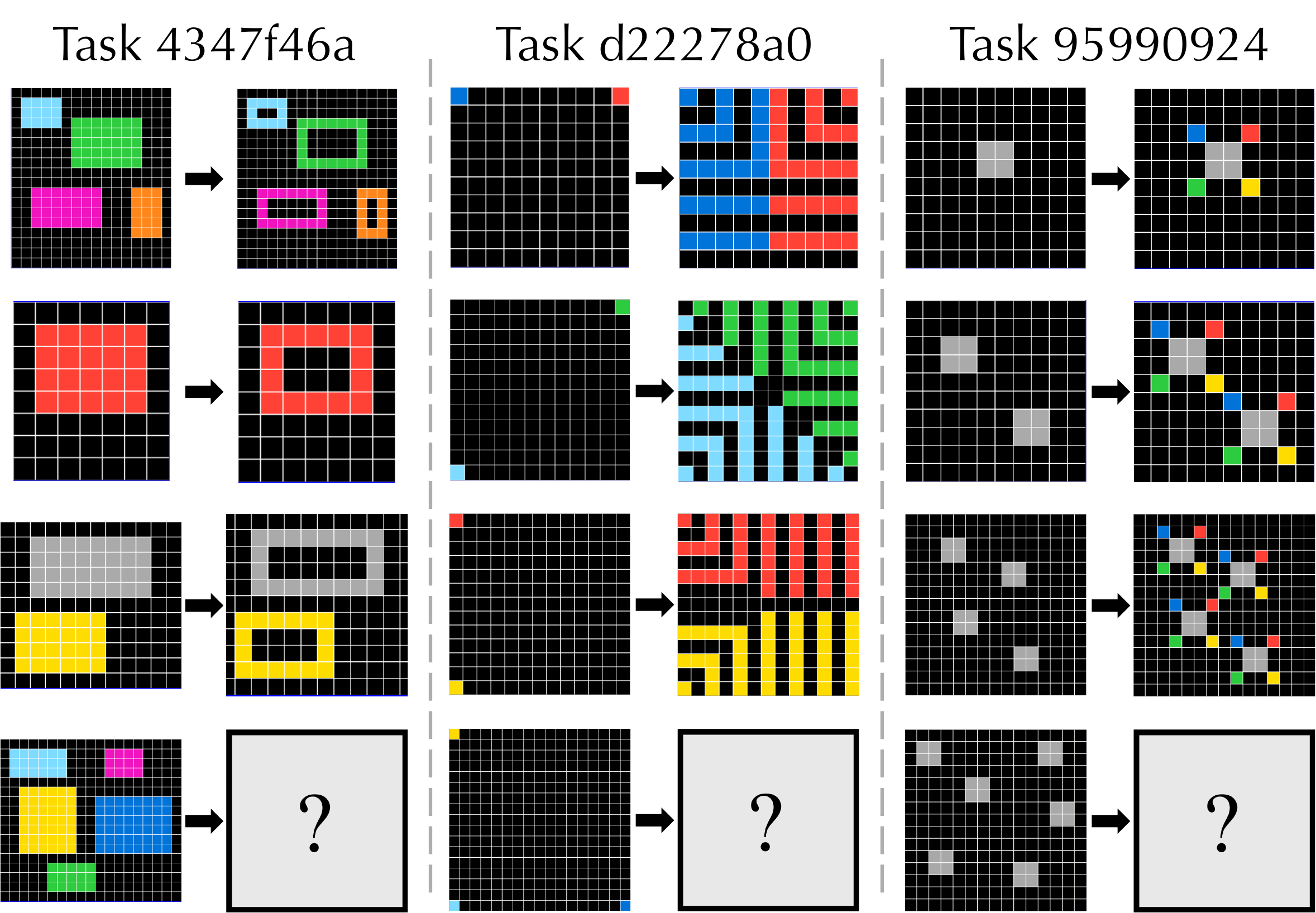}
    \caption{Example tasks from the ARC-AGI-1 dataset.}
    \label{fig:arcexamples}
\end{figure}

The Abstraction and Reasoning Corpus (ARC-AGI)\citep{chollet2019arc} is a benchmark of grid-based reasoning problems designed to probe generalization from a handful of examples rather than pattern recognition at scale. Each task specifies a latent rule, typically an abstract transformation over objects, colours, and spatial relations, that must be inferred from a small set of input-output demonstrations and then applied to a new query input. Example tasks are visualized in Figure~\ref{fig:arcexamples}.

Formally, an ARC-AGI task consists of $K$ demonstration pairs $\{(x^{(k)}_{\text{in}}, x^{(k)}_{\text{out}})\}_{k=1}^K$ and a query input $x^{(q)}_{\text{in}}$. Each grid $x \in \{0, \ldots, 9\}^{H \times W}$ encodes colours as integers, with dimensions up to $30 \times 30$. The goal is to predict the corresponding query output $x^{(q)}_{\text{out}}$ by identifying the underlying transformation that maps $x_{\text{in}} \mapsto x_{\text{out}}$ in the demonstrations.

ARC-AGI is significant because it emphasizes compositional and systematic reasoning: success often requires discovering object structure (connected components, symmetry, repetition), manipulating discrete entities (copy, translate, rotate, reflect), and applying conditional rules that depend on spatial context. Unlike supervised datasets, where the same mapping is learned across many labelled examples, ARC tasks require on-the-fly rule induction from only a few demonstrations, making it a challenging testbed for models that aim to exhibit flexible, human-like generalization.

\subsection{Discrete Diffusion}
\label{sec:discrete_diffusion}

Discrete diffusion models \citep{d3pm2021} define a generative process over a discrete state space $\mathcal{X} = \{1, \dots, S\}$ via a forward corruption process and a learned reverse denoising process.

\paragraph{Forward Process.}
The forward process $q(x_t | x_{t-1})$ injects noise according to a transition matrix $Q_t \in \mathbb{R}^{S \times S}$, such that $q(x_t | x_{t-1}) = x_{t-1} Q_t$ (where $x$ is represented as a one-hot vector). This Markov chain admits a closed-form marginal distribution $q(x_t | x_0) = x_0 \bar{Q}_t$, where $\bar{Q}_t = Q_1 \dots Q_t$.
In this work, we adopt the Masked Diffusion Language Model (MDLM) framework \citep{mdlm2024}, which uses an absorbing state parameterization. The transition matrix $Q_t$ is defined such that tokens either remain unchanged or transition to a special \texttt{[MASK]} token with probability $\beta_t$. Crucially, once a token is masked, it remains masked (absorbing state).

\paragraph{Reverse Process.}
The generative process reverses this corruption by learning a denoising distribution $p_\theta(x_{t-1} | x_t)$. This is typically parameterized by a neural network $\hat{p}_\theta(x_0 | x_t)$ that predicts the clean data distribution from the noisy state. The reverse transition is then derived using the posterior of the forward process:
\begin{equation}
    p_\theta(x_{t-1} | x_t) \propto \sum_{\tilde{x}_0 \in \mathcal{X}} q(x_{t-1} | x_t, \tilde{x}_0) \hat{p}_\theta(\tilde{x}_0 | x_t).
\end{equation}
Recent works such as LLaDA \citep{llada2025} and ARChitects \citep{architects2025} have applied this framework to 2D grids by flattening them into sequences, but they rely on standard 1D architectures that may introduce topological mismatches for spatial reasoning tasks.

\subsection{Composing Diffusion Models}
\label{sec:composition}

In many settings, we have access to multiple pretrained diffusion ``experts'' $\{p^{(m)}_\theta\}_{m=1}^M$ (e.g., trained on different data sources or with different architectures) and wish to combine them at inference time. As specified in \Cref{sec:discrete_diffusion}, each expert induces a reverse transition $p^{(m)}_\theta(x_{t-1}\mid x_t)$. A natural goal is to sample from a composed target distribution built from the experts.

\paragraph{Naive composition without weights.}
Two common ways to combine experts are:
\begin{align}
\textbf{Mixture (OR):}\quad 
p_t^{\mathrm{mix}}(x) &= \frac{1}{M}\sum_{m=1}^M p_t^{(m)}(x),
\label{eq:naive_mixture}
\\
\textbf{Product (AND):}\quad
p_t^{\mathrm{prod}}(x) &\propto \prod_{m=1}^M p_t^{(m)}(x).
\label{eq:naive_product}
\end{align}
The mixture favours samples that are likely under \emph{any} expert, while the product favours samples that are simultaneously likely under \emph{all} experts. In practice, however, experts are rarely equally reliable, motivating methods that introduce weights. Importantly, existing work typically does so using \textbf{a single scalar per expert}, either as mixture coefficients or as global exponents.

SUPERDIFF~\citep{superdiff2024} formalizes composition by specifying a weighted mixture target
\begin{equation}
p_t^{\mathrm{mix}}(x) \;=\; \sum_{m=1}^M \omega_m\, p_t^{(m)}(x),
\qquad \omega_m \ge 0,\ \sum_m \omega_m = 1,
\label{eq:superdiff_mix}
\end{equation}
and deriving the corresponding superposed reverse dynamics. During sampling, SUPERDIFF estimates $\log p_t^{(m)}(x_t)$ along the trajectory and uses these values to compute state-dependent responsibilities; nevertheless, the user-facing composition parameters remain one scalar coefficient per expert ($\{\omega_m\}$, or equivalent global biases/temperatures).
SUPERDIFF also proposes an ``AND''-style construction that selects coefficients to equalize density changes across experts; this again solves for a small set of scalar weights (one per expert) rather than introducing spatially varying notions of expert competence.

Feynman-Kac Correctors (FKC)~\citep{feynmankac2025,skreta2025feynmankac} considers product/geometric-average targets obtained by adding global exponents to the naive product in \eqref{eq:naive_product}, e.g.
\begin{equation}
p_t^{\mathrm{geo}}(x) \propto \prod_{m=1}^M \big(p_t^{(m)}(x)\big)^{\beta_m},
\label{eq:fkc_geo}
\end{equation}
FKC derives a Feynman--Kac reweighting that corrects heuristic score-mixing so that a particle system tracks the intended intermediate targets. While the algorithm uses SMC importance weights $w_t^{(j)}$ to correct samples, these weights are scalar per particle, and the composition itself is governed by scalar per-expert exponents $\{\beta_m\}$, fixed across the state space.

RNE~\citep{rne2025} similarly targets annealed and product-style compositions by introducing global exponents on expert densities, e.g.
\begin{equation}
p_0(x) \propto \prod_{m=1}^M \big(p_0^{(m)}(x)\big)^{\beta_m},
\label{eq:rne_prod}
\end{equation}
and uses Radon--Nikodym / time-reversal identities to estimate marginal density ratios along trajectories. As in FKC, the resulting composition remains parameterized by one scalar per expert through $\{\beta_m\}$.


\section{Factor-Wise Composition}
\label{sec:method}

The correction methods reviewed in \cref{sec:composition}, such as Feynman-Kac correctors \citep{feynmankac2025}, RNE \citep{rne2025}, and weighted score addition \citep{superdiff2024}, are \emph{target-distribution-first}: they specify a compositional target distribution, typically the product of experts $p(x) \propto \prod_m p_m(x)$, and reverse-engineer a sampling procedure to match it through modified rate matrices, importance weights, and particle resampling. In essentially all cases, composition is parameterized by one scalar weight per expert, for example, through a weighted product:
\begin{equation}
p_{\mathrm{comp}}(x) \;\propto\; \prod_{m=1}^{M} \big(p_m(x)\big)^{\alpha_m},
\qquad \alpha_m \ge 0.
\label{eq:global_expert_weights}
\end{equation}
This formulation assumes that every expert provides a globally competent distribution over the full state: each expert's predictions over \emph{all} variables are always folded into the product at every position, if an expert is up-weighted, it is up-weighted everywhere in $x$. Importance weighting can correct for discrepancies between the sampling process and the target distribution, but it cannot correct for a misspecified target.

In structured domains, however, the state often admits a natural decomposition into factors (e.g., spatial regions, variable groups, constraint scopes, or output components), as is standard in probabilistic graphical models~\citep{pgmbook}. When experts are complementary specialists, global weights in \cref{eq:global_expert_weights} can be fundamentally misspecified: a shape expert trained only on occupancy produces arbitrary predictions over colour tokens, yet the product includes them at every position; a colour expert that hallucinates shapes sees those hallucinations enter the product alongside the shape expert's accurate predictions. The desired composition is to trust the shape expert for shape and the colour expert for colour. However, this cannot be expressed as $\prod_k p_k^{\alpha_k}$ for any choice of global exponents $\alpha_k$; it requires composition decisions that vary across the state space. Moreover, even if such a position-dependent target could be specified, existing methods can only correct at the sample level (i.e., using singular scalar weights per trajectory) and cannot recover the combinatorial space of per-position expert assignments (\cref{sec:composition}).

This suggests replacing global scalar weighting with \emph{factor-wise} composition, where each factor $f \in \mathcal{F}$ can be assigned its own expert weighting:
\begin{equation}
p_{\mathrm{comp}}(x) \;\propto\; \prod_{f \in \mathcal{F}} \ \prod_{m=1}^{M} \big(p_m(x_f)\big)^{\alpha_{m,f}}.
\label{eq:factorwise_expert_weights}
\end{equation}
Here $x_f$ denotes the subset/view of variables belonging to factor $f$, and the weights $\{\alpha_{m,f}\}$ allow expertise to vary across factors. In the hard-routing special case, $\alpha_{m,f}$ is one-hot in $m$, yielding a single selected expert per factor.

FactorDiff operationalizes this factor-wise perspective during inference. At each denoising step, a routing map assigns each variable (position/pixel) to the most competent expert. This defines a valid generative process that respects expert specialization without committing to a product-of-experts target, and enables composition decisions at the spatial/positional level.

\subsection{Composition as Routing}
\label{sec:routing_framework}

Masked diffusion models such as MDLM \citep{mdlm2024} factorize the reverse transition over variables $v \in V$:
\begin{equation}
    p_\theta(x_{t-1} \mid x_t) = \prod_{v \in V} p_\theta(x_{t-1,v} \mid x_t).
    \label{eq:factorized_reverse}
\end{equation}
We exploit this factorization to compose experts at the variable level. Given $M$ expert models, we introduce a routing variable $z_v \in \{1, \dots, M\}$ that assigns each variable $v$ to a specific expert. The composed transition is:
\begin{equation}
    p(x_{t-1} \mid x_t, z) = \prod_{v \in V} p_{z_v}(x_{t-1, v} \mid x_t),
    \label{eq:routed_transition}
\end{equation}
where $p_{z_v}(x_{t-1, v} \mid x_t)$ is the per-variable conditional predicted by expert $z_v$. Since each factor is a valid conditional from a trained model, the product defines a valid joint transition; no additional normalization or importance correction is required.

The expressiveness of this framework is governed by the granularity of the routing map $z$. We identify two levels:

\paragraph{Per-Sample Composition ($z_v = z$ for all $v$).}
When the routing map is constant across positions, composition can only select among complete samples. Existing importance-weighted methods operate in this regime. Feynman-Kac correctors \citep{feynmankac2025} and RNE \citep{rne2025} combine expert predictions identically at every position, then select or discard complete trajectories via scalar importance weights. All positions within a selected sample share the same provenance, and expert contributions cannot vary spatially. Recovering factored compositions (e.g., expert~1 at boundaries, expert~2 at interiors) requires the particle population to cover a combinatorial space that grows exponentially in the number of independently varying positions.

\begin{figure}[t]
    \centering
    \includegraphics[width=\columnwidth]{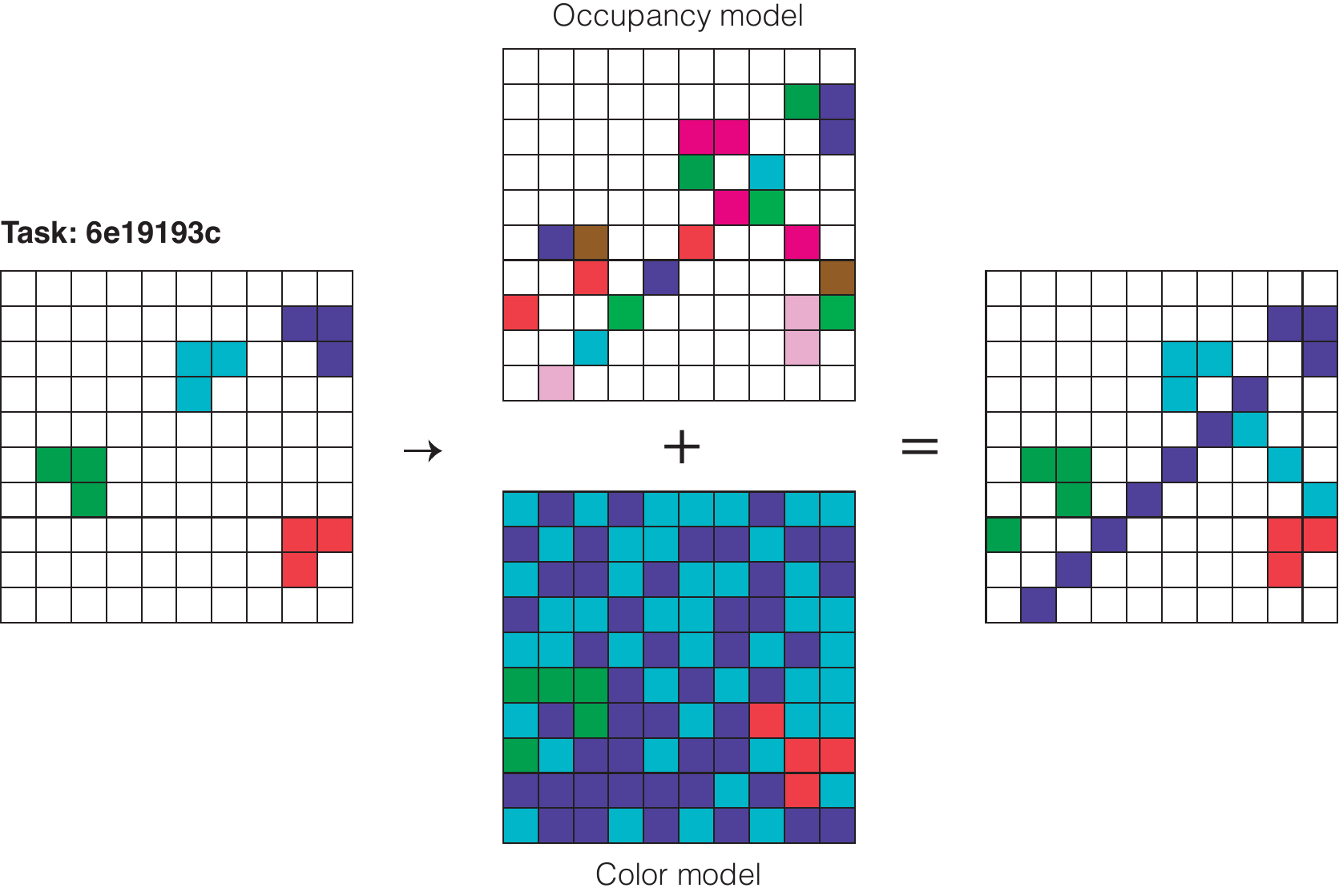}
    \caption{The output space is split: the Occupancy model handles background and borders, while the Colour model fills in the texture. This level of factorization is impossible at the sample-level.}
    \label{fig:perpixel}
\end{figure}

\paragraph{Per-Pixel Composition ($z_v$ varies across $v$).}
The focus of this work is to explore the potential of allowing the routing map to vary across positions (pixels). This enables compositions that no single sample-level particle can express: one expert's confident boundary predictions stitched together with another's confident interior predictions within a single sample (\cref{fig:perpixel}). We show that position-level routing handles all tested expert combinations: pairs of full models, a spatial expert with a full model, and two different expert models, consistently and without modification. We instantiate this via confidence-based selection in \cref{sec:perpixel}.

\subsection{Confidence-Based Routing}
\label{sec:perpixel}

\begin{figure*}[t]
    \centering
    \includegraphics[width=0.68\textwidth]{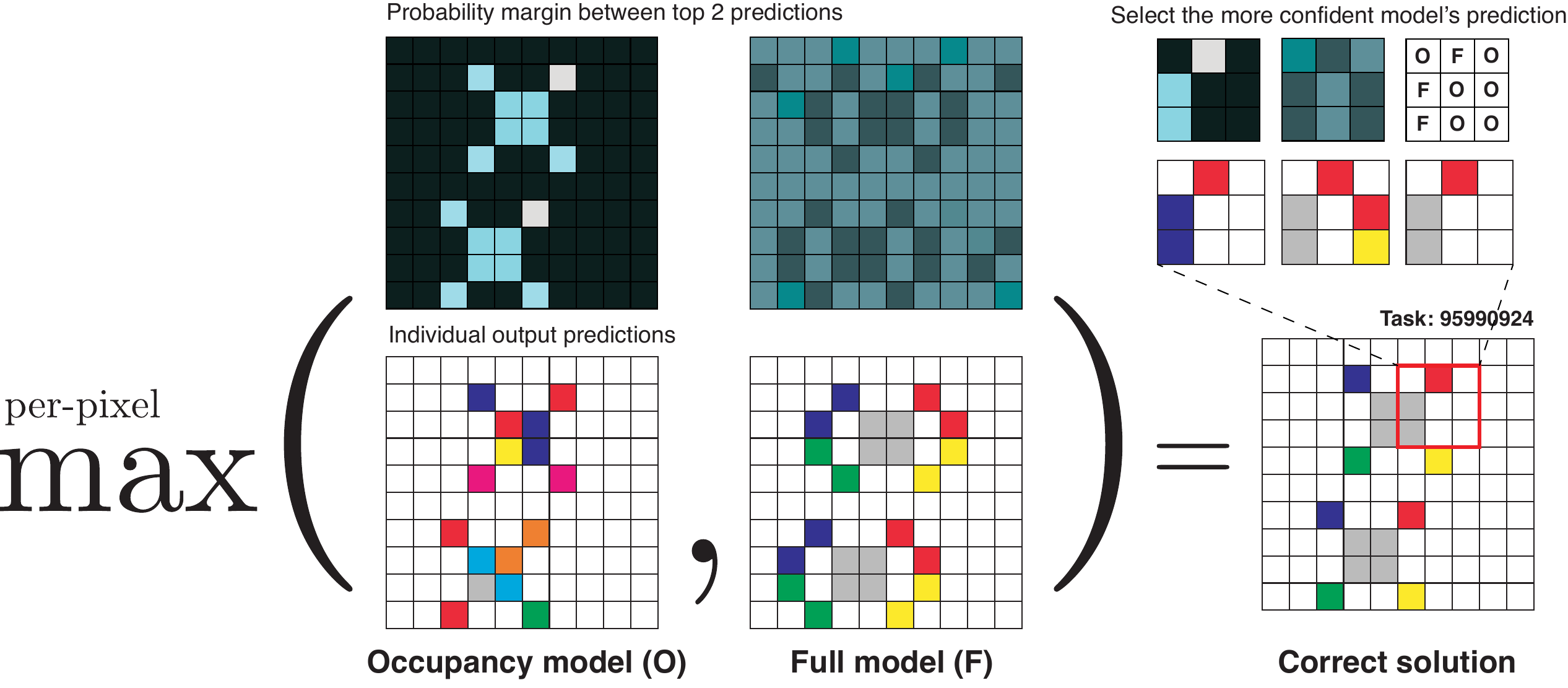}
    \caption{Per-pixel routing in action. The routing map $z$ (visualized as a mask) is computed at each denoising step. The Occupancy model is selected at boundaries where it is confident, providing sharp shapes. The Full model is selected for interiors, providing colour.}
    \label{fig:pixelmax}
\end{figure*}

When expertise varies spatially (e.g., one Occupancy model excels at boundaries while another Colour model excels at textures), we employ spatial routing. In this work, we explore a simple instantiation of this concept, in which the routing variable $z_v$ is dynamically determined at inference time based on expert confidence (\cref{fig:pixelmax}).

We select the expert with the highest confidence at each position $v$:
\begin{equation}
    z_v = \operatorname*{argmax}_{k \in \{1,\dots,K\}} \mathcal{C}(p_k(x_v \mid x_t)),
    \label{eq:pixelmax}
\end{equation}
where the confidence metric $\mathcal{C}$ is the margin between the top two probabilities, as inspired by the work of \citet{tokenordering2025}:
$\mathcal{C}(p) = p^{(1)} - p^{(2)}$. The composed reverse step is then:
\begin{equation}
    p_{\text{composed}}(x_{t-1}^v \mid x_t) = \sum_{k=1}^K \mathbb{I}[z_v = k] \cdot p_k(x_{t-1}^v \mid x_t).
\end{equation}

\section{Problem Formulation}



Standard approaches flatten grids into sequences $x \in \mathbb{R}^{HW}$, imposing an arbitrary ordering (raster scan) that forces attention to learn that tokens $i$ and $i+W$ are vertical neighbours. We instead treat the grid as a lattice graph $G = (V, E)$, where $V$ is the set of cells and $E$ encodes adjacency. Specializing the general discrete diffusion framework of \cref{sec:discrete_diffusion} to this 2D setting, each variable $v \in V$ corresponds to a grid cell taking values in $\mathcal{X} = \{0, \ldots, 9\}$ (10 colours for ARC-AGI) plus a \texttt{[MASK]} token.


\subsection{ARC-AGI-1}
We train on a subset of ARC-AGI-1 tasks. Specifically, we select 120 tasks in which grids fit within a $10 \times 10$ grid, covering colour permutations, object detection, symmetry, translations, and pattern completion (see \cref{sec:appendix_samples} for details). We then use the RE-ARC generator \citep{re_arc2024} to generate 1000 unique instances for each task, yielding 120,000 total example pairs. Of the 1000 generated instances per task, we train on 800 and evaluate in-distribution on the held-out 200. We apply geometric augmentations (90$^\circ$ rotations, horizontal and vertical flips, and their compositions), yielding up to 8 equivalent views per training example.


Rather than learning explicit task embeddings, we provide the model with \textbf{context frames}: input-output demonstration pairs concatenated along the channel dimension with the query input. This enables a single model to be trained on all 120 tasks simultaneously, in contrast to prior work that trains separate models per task~\citep{li2025tackling}. During inference, the model identifies which transformation to apply purely from the provided context frames.

\paragraph{Expert Variants.}
We train three complementary model types, each with a different training objective (see \cref{sec:appendix_loss} for detailed equations):
\begin{itemize}
    \item \textbf{Full model}: Standard masked cross-entropy over all 10 colors. Learns the complete input-to-output mapping.
    \item \textbf{Occupancy model (Occ)}: Binary cross-entropy for background (colour 0) versus foreground (colours 1--9). Learns where objects appear without distinguishing between colours.
    \item \textbf{Color model (Col)}: Cross-entropy computed only at foreground positions (colors 1--9), ignoring background. Focuses on predicting which colour appears at non-background positions.
\end{itemize}





\subsection{Masked Diffusion on Grids}
\label{sec:masked_diffusion}

\begin{figure*}[t]
    \centering
    \includegraphics[width=0.58\textwidth]{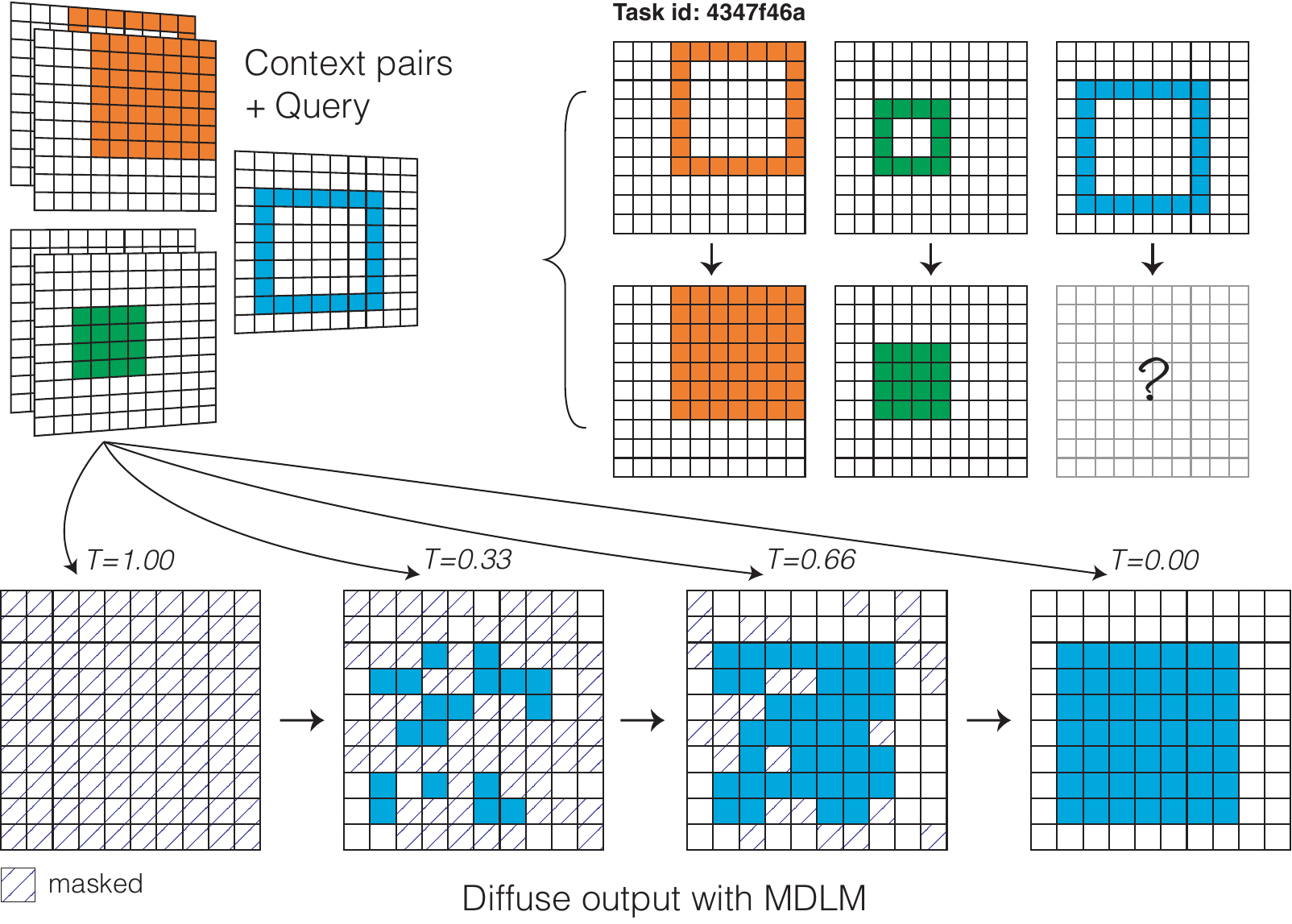}
    \caption{Masked diffusion on 2D grids. The forward process progressively masks tokens (left to right), while the reverse process iteratively unmasks them conditioned on the input grid, recovering the clean output.}
    \label{fig:mdlm}
\end{figure*}

We employ continuous-time masked diffusion \citep{mdlm2024}, specialized to 2D grids (\cref{fig:mdlm}). Starting from a fully-masked state (maximum entropy), the reverse process progressively reveals tokens according to a noise schedule.


Following \citet{mdlm2024}, we enforce two properties via the SUBS parameterization: (1) the output never predicts \texttt{[MASK]}, implemented by setting $\text{logits}[\texttt{MASK}] = -\infty$; and (2) unmasked tokens copy themselves with probability 1. The continuous-time NELBO reduces to weighted masked cross-entropy:
\begin{equation}
\mathcal{L} = \mathbb{E}_{t, q(x_t|x_0)} \left[ \frac{\sigma'_t}{1 - \alpha_t} \cdot \text{CE}(p_\theta(x_t), x_0) \right]
\end{equation}
where $\sigma'_t = d\sigma/dt$ and the loss is computed only at masked positions.

\subsection{Denoising Architecture}
\label{sec:architecture}

The masked diffusion process described in \cref{sec:masked_diffusion} defines the outer generative loop: starting from a fully masked grid, it progressively unmasks tokens over $T$ denoising steps. At each step $t$, a neural network must predict the clean data distribution $\hat{p}_\theta(x_0 \mid x_t)$. We parameterize this network using an iterative refinement architecture, which serves as the denoising model for all expert distributions $p_k(x_{t-1}^v \mid x_t)$ defined in \cref{eq:routed_transition}.

\paragraph{Iterative Refinement Backbone.}
Following the iterative refinement paradigm of TRM \citep{trm2024}, the denoising network maintains dual latent states $z_H$ (high-level) and $z_L$ (low-level). Within a single denoising step at time $t$, the network performs $H$ internal refinement cycles, allowing constraint propagation across the grid before producing the final prediction. Learnable 2D positional embeddings are added to token embeddings, ensuring that the confidence metric $\mathcal{C}(p_k)$ in \cref{eq:pixelmax} reflects true spatial locality rather than artifacts of raster scanning.

\paragraph{Backbone Variants.}
We implement two variants for $f_{\text{backbone}}$, both conditioned on the query input through channel-wise concatenation:
\begin{itemize}
    \item \textbf{Convolutional (Conv-IR)}: A U-shaped encoder-decoder with skip connections. Local receptive fields capture fine-grained spatial patterns, while hierarchical downsampling aggregates global context.
    \item \textbf{Transformer (Trans-IR)}: A bidirectional attention-based architecture where each token attends to all others, providing immediate global context, with 2D sinusoidal positional embeddings to preserve grid topology.
\end{itemize}

\section{Experimental Results}
\label{sec:basemodel_experiments}

Our experiments are designed to answer two questions:
\begin{itemize}
    \item \textbf{RQ1:} Does per-pixel routing outperform per-sample composition when experts are specialized?
    \item \textbf{RQ2:} Is per-pixel routing ``safe'' when experts are both strong generalists (full+full), i.e., does it avoid degrading performance relative to the best available baseline?
\end{itemize}

\paragraph{Metrics.}
We report \textit{Pixel accuracy}, the fraction of correctly predicted cells across all grids, and \textit{Exact accuracy}, the fraction of grids for which every cell is correct. All results are evaluated on the same 120-task ARC-AGI-1 subset, using RE-ARC-generated held-out examples (IID evaluation).


\paragraph{Single-expert baselines.}
\Cref{tab:basemodel_results} reports the performance of individual experts on the held-out RE-ARC evaluation set. The Trans-IR full model achieves 93.0\% exact accuracy, outperforming the Conv-IR full model (89.3\%). In contrast, the specialist experts perform poorly on the full 10-class end task (3.3\% exact for the colour expert, 0.2\% exact for the occupancy expert), since they are trained on restricted objectives. This gap motivates composition: specialists may be inaccurate globally, but highly reliable on the substructure they are intended to model.




\begin{table}[ht]
    \centering
    \caption{Single expert performance on ARC-AGI-1 tasks}
    \label{tab:basemodel_results}
    \begin{tabular}{llccc}
        \toprule
        Backbone & Variant & Params & Pixel (\%) & Exact (\%) \\
        \midrule
        Conv-IR & Full & 1M & 99.3 & 89.3 \\
        Trans-IR & Full & 1M & 99.5 & 93.0 \\
        \midrule
        Trans-IR & Col & 1M & 32.4 & 3.3 \\
        Trans-IR & Occ & 1M & 65.7 & 0.2 \\
        \bottomrule
    \end{tabular}
\end{table}


\paragraph{Composition protocols.}
To isolate the effect of routing granularity, we compare \textbf{per-sample} composition methods (PoE averaging, SuperDiff, FKC, RNE) against \textbf{per-pixel} routing (FactorDiff maximum selection). Results are summarized in \Cref{tab:arc_composition_results}.


\subsection{RQ1: Benefit under Specialization (Occ + Full; Col + Occ)}
We evaluate regimes where experts are explicitly complementary, and per-sample methods must use a single global mixture/exponent across all positions.

\textbf{Occ + Full.}
Per-pixel routing achieves 94.0\% exact accuracy, outperforming the strongest per-sample method (SuperDiff density at 93.2\%) by 0.8\%. More importantly, several per-sample methods \emph{degrade} relative to the best single expert (93.0\%): PoE average drops to 91.0\% and FKC to 90.2\%, consistent with global composition diluting correct full-model predictions in regions where the occupancy expert is uninformative. Per-pixel routing avoids this trade-off by selecting the occupancy expert only where it is locally reliable, while deferring elsewhere.

\textbf{Col + Occ.}
This pairing directly stresses RQ1: both specialists are nearly useless on the end task individually (3.3\% and 0.2\% exact), yet their errors are structured and complementary. Per-pixel routing composes them into a strong solver, reaching 90.5\% exact accuracy. In contrast, all per-sample methods remain far below (73.2--78.4\%), with a 12+ point gap to the best per-sample baseline (FKC at 78.4\%). This gap is consistent with the ``scalar bottleneck'': a single global mixture/exponent cannot express ``use occupancy for shape/boundaries and colour for texture'' within the same sample, whereas per-pixel routing can.

\subsection{RQ2: Safety under Strong Generalists (Full + Full)}
We test whether per-pixel routing harms performance when both experts are competent across the full state space. Composing two strong full models (Trans-IR 93.0\% and Conv-IR 89.3\%), all composition methods improve over the individual baselines. Crucially, per-pixel routing is \emph{non-degrading}: it matches the best per-sample baseline, tying FKC at 95.2\% exact accuracy (and matching or exceeding PoE average and SuperDiff density at 95.1\%). This indicates that allowing spatially varying expert selection does not introduce systematic stitching errors or instability when experts already agree broadly; instead, it behaves like a benign refinement that preserves strong global competence. 

Across ARC-AGI-1, per-pixel routing is \emph{safe} in the generalist regime (Full+Full; RQ2) and becomes increasingly advantageous as expert specialization increases (Occ+Full and especially Col+Occ; RQ1). These results support the central claim that factor-wise composition can capture complementary expertise that per-sample/global-weight composition cannot represent. 

\begin{table*}[t]
    \centering
    \caption{Composition results on 120-task ARC-AGI-1 subset (RE-ARC IID evaluation). We compare composition methods across expert combinations. Per-pixel max consistently outperforms or matches all baselines across all combinations, including col+occ where it dramatically exceeds alternatives.}
    \label{tab:arc_composition_results}
    \resizebox{0.535\linewidth}{!}{
    \begin{tabular}{llccc}
        \toprule
        \multirow{2}{*}{Granularity} & \multirow{2}{*}{Method} & Full + Full & Occ + Full & Col + Occ \\
        \cmidrule(lr){3-3} \cmidrule(lr){4-4} \cmidrule(lr){5-5}
        & & Exact (\%) & Exact (\%) & Exact (\%) \\
        \midrule
        \multicolumn{2}{l}{\textit{Single model baselines}} \\
        --- & Full (Trans-IR, 93.0\%) & 93.0 & 93.0 & --- \\
        --- & Full (Conv-IR, 89.3\%) & 89.3 & --- & --- \\
        --- & Occ & --- & 0.2 & 0.2 \\
        --- & Col & --- & --- & 3.3 \\
        \midrule
        \multicolumn{2}{l}{\textit{Per-sample routing}} \\
        Per-sample & Average (PoE) & 95.1 & 91.0 & 73.2 \\
        Per-sample & Density (SuperDiff) & 95.1 & 93.2 & 77.7 \\
        Per-sample & FKC & \textbf{95.2} & 90.2 & 78.4 \\
        Per-sample & RNE & 94.1 & 85.2 & 76.5 \\
        \midrule
        \multicolumn{2}{l}{\textit{Per-pixel routing}} \\
        Per-pixel & Maximum selection & \textbf{95.2} & \textbf{94.0} & \textbf{90.5} \\
        \bottomrule
    \end{tabular}
    }
\end{table*}

\subsection{Expert-Overlap Ablation and Routing Maps}
\label{sec:overlap}
\label{sec:appendix_overlap}

\begin{figure*}[ht]
    \centering
    \includegraphics[width=0.9\linewidth]{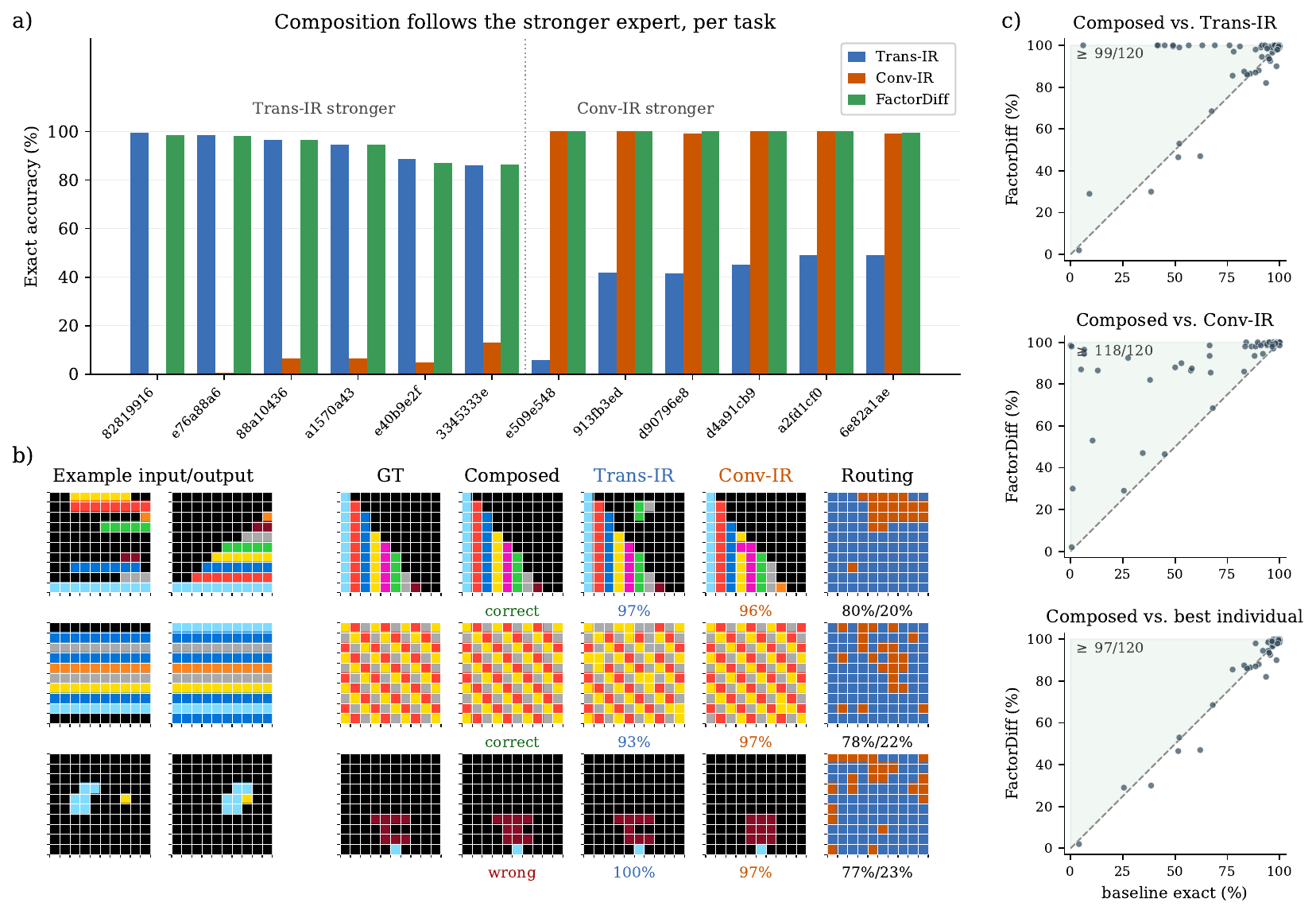}
    \caption{\textbf{Expert overlap ablation and routing maps} (Trans-IR vs.\ Conv-IR: same data and objective, differ only in architecture; 120 tasks, 24{,}000 samples, untuned). \textbf{a)} On the 12 most-disagreeing tasks, composition follows the stronger expert in both directions. \textbf{b)} Routing maps: top two rows are correct where neither model is, by stitching each model's correct regions; bottom row is a failure case. \textbf{c)} Per task, composition matches or exceeds each single model (Trans-IR 99/120, Conv-IR 118/120) and is not bounded by the better one.}
    \label{fig:overlap_full}
\end{figure*}

In \cref{fig:overlap_full}, we inspect behaviour when experts overlap heavily; we run a Full+Full ablation study, in which our composite FactorDiff framework is applied over two full-output experts trained with the same objective and data distribution but differing only in architectural inductive bias (Trans-IR and Conv-IR). We evaluate all 200 held-out queries for each of the 120 tasks (24{,}000 samples). This pair is separate from the main aggregate in \cref{tab:arc_composition_results}, which uses the stronger headline full models; here the goal is to probe routing behaviour under heavy overlap. Per-pixel routing achieves 94.8\% exact accuracy, well above both deployable models (89.5\% and 85.6\%). Crucially, for 132 of the 24{,}000 grids, the composed output is exactly correct, whereas neither individual model is. Per-pixel routing stitches correct regions from each into solutions neither produces alone. Even with hindsight, the better of the two models chosen per task, composition wins on more tasks than it trails (39 vs 23, with 58 ties), without ever being catastrophically worse. In the figure, we also show bar charts and scatter plots that provide further insight into per-task performance of compositional models vs single models.



\section{Related Works}

\paragraph{Existing ARC-AGI Solvers.}

Early ARC-AGI solving approaches focused on symbolic methods~\citep{wind2020kaggle,alford2021dream,xu2023graphs,hocquette2025relational}, but recent work has shifted towards latent methods leveraging language or vision models~\citep{franzen2025product,li2025tackling,hu2026arc}. While early LLMs struggled (i.e., GPT-4 solved only 13/50 simple tasks~\citep{xu2024llms}), newer models achieve much higher performance, with GPT-5.5 achieving 95\%. More recently,
TRM \citep{trm2024} achieved 45\% with 7M parameters via latent recursion, though critiques suggest heavy reliance on search rather than generalizable reasoning \citep{trm_analysis2024}. CompressARC \citep{compressarc2025} and SOAR \citep{soar2025} take a program synthesis approach and search for code reproducing examples, but suffer from combinatorial explosion and brittleness. Compositional-ARC \citep{compositional_arc2025} tests recombination of learned primitives, but operates at the task level: composing entire sub-programs rather than expert predictions within a single generative process. In contrast, FactorDiff composes expert predictions at the position level during diffusion sampling, enabling fine-grained spatial routing without requiring an explicit program decomposition.

\paragraph{Factor-based Composition for Continuous Diffusion.} The idea of composing diffusion models at the factor level has been explored in the continuous setting. These methods compose factors by adding their scores over the graph and apply Bethe-style corrections to avoid double-counting variables shared across overlapping scopes \citep{yang2023compositional,xu2024set,sridhar2024adapting,mishra2023GSC,mishra2024GFC,diffcollage}. Our work brings this factor-level perspective to discrete diffusion: rather than adding continuous scores over a fixed factor-graph topology, it routes each factor to a single expert, which composes valid conditionals directly and requires no correction for overlap.

\paragraph{Model Merging and Mixtures.}
A broad literature merges pre-trained models by averaging weights (model soups~\citep{wortsman2022soups}) or routing inputs to jointly-trained sub-networks (mixture-of-experts~\citep{shazeer2017moe}). We instead combine frozen experts purely at inference: there is no shared parameter space, so weight-space merging does not apply, and unlike prior diffusion composition that mixes full outputs through one scalar weight per expert (\cref{sec:composition}), FactorDiff mixes experts per factor within intermediate denoising states. This gives a finer-grained composition regime than scalar output-level fusion.

\section{Conclusion}
\label{sec:conclusion}

We introduced FactorDiff, a factor-wise composition framework for discrete diffusion models that operates at the position level rather than the sample level. Built on a native 2D masked diffusion architecture that preserves grid topology, FactorDiff dynamically routes each position to the most competent expert via confidence-based selection. The key insight is that existing per-sample composition methods, which assign a single scalar weight to each expert, cannot capture the spatially varying expertise structure of complementary expert models.

On RE-ARC (120 tasks, IID evaluation), FactorDiff's per-pixel routing consistently achieves the best performance across expert combinations. It further achieves 90.5\% exact accuracy by composing colour and occupancy specialists that individually score only 3.3\% and 0.2\% on the full task, dramatically exceeding per-sample methods (best: 78.4\%). 

Overall, FactorDiff provides a simple and effective mechanism for combining arbitrary specialists within a single diffusion trajectory, and naturally extends to learned or task-conditioned routing when expertise structure is unknown.

\paragraph{Limitations and Open Questions}
\label{sec:limitations}

Our confidence-based routing assumes that expert confidence correlates with correctness, which holds across all tested combinations but may not generalize to arbitrary model pairs (e.g., architectures with unknown complementary strengths). A natural extension is to \emph{learn} the routing function (e.g., train a small network to predict per-pixel routing weights from the current diffusion state).
This would generalize FactorDiff to settings where expertise structure must be discovered rather than designed.

We instantiate factors as individual pixels, but this is one realization of a more general principle: factors could equally be objects, connected regions, or latent abstractions, and discovering informative factorizations automatically 
is itself non-trivial. We note, however, that manually designed factorizations are not strictly necessary to benefit from per-pixel mixing, and with a well-designed routing function, composing two full-output experts already improves over the best individual in our overlap ablation (\cref{sec:overlap}). We leave richer factorizations to future work.

We establish that fine-grained, per-factor routing for discrete diffusion models is feasible and consistently beneficial in an unstudied regime, without proving the optimality of confidence-based routing. Characterizing when factor-wise composition provably helps, and extending beyond ARC-AGI, are natural next steps.


\clearpage

\begin{acknowledgements}
A.A.-G. thanks Anders G. Fr{\o}seth for his generous support. A.A.-G. also acknowledges the generous support of Natural Resources Canada and the Canada 150 Research Chairs program. 

\acknowSciNet{}
\end{acknowledgements}

\bibliography{references}

\newpage
\appendix
\onecolumn

\title{From Global to Factor-Wise Expert Composition in Discrete Diffusion Models\\(Supplementary Material)}

\maketitle


\section{Supplementary Material}
\label{sec:appendix}




\subsection{Transformer Backbone Architecture}
\label{sec:appendix_architecture}

This section provides implementation details for the Trans-IR backbone described in \cref{sec:architecture}.

\paragraph{Input Embedding.}
The input grid $x_t \in \{0, \ldots, 10\}^{H \times W}$ (10 colors plus mask token) is embedded via a learnable embedding table $\mathbf{E} \in \mathbb{R}^{11 \times d}$, producing token embeddings of dimension $d$. The query input $c = x^{(q)}_{\text{in}}$ is embedded separately and concatenated channel-wise, doubling the embedding dimension to $2d$ before a linear projection back to $d$:
\begin{equation}
\mathbf{h}^{(0)} = \mathbf{W}_{\text{in}} [\mathbf{E}(x_t) \| \mathbf{E}(c)] + \text{PE}
\end{equation}
where $\mathbf{W}_{\text{in}} \in \mathbb{R}^{d \times 2d}$ and PE denotes positional encodings.

\paragraph{2D Positional Encoding.}
We use 2D sinusoidal positional encodings to preserve grid topology. For position $(i, j)$, we compute separate encodings for each spatial dimension and concatenate them:
\begin{equation}
\text{PE}_{(i,j)} = [\sin(\omega_k i), \cos(\omega_k i), \sin(\omega_k j), \cos(\omega_k j)]_{k=1}^{d/4}
\end{equation}
where $\omega_k = 1 / 10000^{4k/d}$. This produces a $d$-dimensional encoding that captures both row and column position independently.

\paragraph{Context Conditioning.}
Rather than conditioning on the noise level $\sigma_t$ directly, we condition on context embeddings derived from the input. The query input $c$ is encoded via a small convolutional network followed by global average pooling:
\begin{equation}
\mathbf{c}_{\text{ctx}} = \text{MLP}(\text{Pool}(\text{ConvNet}(\text{onehot}(c))))
\end{equation}
When context pairs $(x^{(k)}_{\text{in}}, x^{(k)}_{\text{out}})$ are provided, they are similarly encoded and averaged, then added to the context embedding.

\paragraph{Adaptive Layer Normalization (adaLN-Zero).}
Each transformer block is modulated by the context embedding via adaLN-Zero, which produces six modulation parameters:
\begin{equation}
(\gamma_1, \beta_1, \alpha_1, \gamma_2, \beta_2, \alpha_2) = \text{MLP}(\mathbf{c}_{\text{ctx}})
\end{equation}
The modulation is applied as:
\begin{equation}
\text{adaLN}(\mathbf{h}, \gamma, \beta) = \gamma \odot \text{LayerNorm}(\mathbf{h}) + \beta
\end{equation}
where $\gamma$ and $\beta$ are broadcast over the sequence dimension. The $\alpha$ parameters scale the residual connections, initialized to zero for stable training.

\paragraph{Transformer Block.}
Each transformer block consists of:
\begin{enumerate}
    \item \textbf{Self-attention}: Multi-head attention with Flash Attention over all $H \times W$ tokens
    \item \textbf{Feed-forward}: Two-layer MLP with GELU activation
\end{enumerate}

The forward pass for block $\ell$ with adaLN-Zero modulation is:
\begin{align}
\mathbf{h}' &= \mathbf{h}^{(\ell)} + \alpha_1 \odot \text{Attn}(\text{adaLN}(\mathbf{h}^{(\ell)}, \gamma_1, \beta_1)) \\
\mathbf{h}^{(\ell+1)} &= \mathbf{h}' + \alpha_2 \odot \text{FFN}(\text{adaLN}(\mathbf{h}', \gamma_2, \beta_2))
\end{align}

\paragraph{Output Projection.}
The final hidden states are layer-normalized and projected to logits over the vocabulary:
\begin{equation}
\text{logits} = \text{LayerNorm}(\mathbf{h}^{(L)}) \mathbf{W}_{\text{out}} + \mathbf{b}_{\text{out}} \in \mathbb{R}^{H \times W \times 11}
\end{equation}
Following SUBS parameterization, we set $\text{logits}[\texttt{MASK}] = -\infty$ to prevent predicting the mask token.

\paragraph{Model Configurations.}
\Cref{tab:model_config} lists the hyperparameters for our transformer backbone.

\begin{table}[ht]
    \centering
    \caption{Transformer backbone configuration ($\sim$1M parameters).}
    \label{tab:model_config}
    \begin{tabular}{lc}
        \toprule
        Hyperparameter & Value \\
        \midrule
        Hidden dimension $d$ & 144 \\
        Number of blocks $L$ & 2 \\
        Attention heads & 4 \\
        Head dimension & 36 \\
        FFN expansion & 4$\times$ \\
        Context dimension & 144 \\
        \bottomrule
    \end{tabular}
\end{table}

\subsection{Convolutional Backbone Architecture}
\label{sec:appendix_conv}

This section provides implementation details for the Conv-IR backbone described in \cref{sec:architecture}.

\paragraph{Input Encoding.}
The input grid $x_t$ and query input $c$ are converted to one-hot representations and concatenated along the channel dimension, producing a tensor of shape $(B, 2V, H, W)$ where $V = 10$ is the vocabulary size. A convolutional layer projects this to the model channel dimension:
\begin{equation}
\mathbf{h}^{(0)} = \text{Conv}_{3 \times 3}([\text{onehot}(x_t) \| \text{onehot}(c)]) \in \mathbb{R}^{B \times C \times H \times W}
\end{equation}

\paragraph{Residual Block.}
Each residual block consists of two $3 \times 3$ convolutions with GroupNorm and SiLU activation, plus context injection:
\begin{align}
\mathbf{h}_1 &= \text{Conv}_{3 \times 3}(\text{SiLU}(\text{GroupNorm}(\mathbf{h}))) \\
\mathbf{h}_2 &= \mathbf{h}_1 + \text{MLP}(\mathbf{c}_{\text{ctx}})_{[:,:,\text{None},\text{None}]} \\
\mathbf{h}' &= \mathbf{h} + \text{Conv}_{3 \times 3}(\text{SiLU}(\text{GroupNorm}(\mathbf{h}_2)))
\end{align}
The context embedding $\mathbf{c}_{\text{ctx}}$ is broadcast spatially and added after the first convolution, allowing the model to modulate its behavior based on the task context.

\paragraph{Context Conditioning.}
Context conditioning follows the same approach as the transformer backbone: a small convolutional network encodes the query input, and optional context pairs are encoded and averaged:
\begin{equation}
\mathbf{c}_{\text{ctx}} = \text{MLP}(\text{Pool}(\text{ConvNet}(\text{onehot}(c)))) + \frac{1}{K} \sum_{k=1}^{K} \text{PairEnc}(x^{(k)}_{\text{in}}, x^{(k)}_{\text{out}})
\end{equation}

\paragraph{Output Projection.}
The final hidden states are normalized and projected to logits:
\begin{equation}
\text{logits} = \text{Conv}_{3 \times 3}(\text{SiLU}(\text{GroupNorm}(\mathbf{h}^{(L)}))) \in \mathbb{R}^{B \times 11 \times H \times W}
\end{equation}
The output is permuted to $(B, H \cdot W, 11)$ for compatibility with the diffusion loss.

\paragraph{Model Configurations.}
\Cref{tab:conv_config} lists the hyperparameters for our convolutional backbone.

\begin{table}[ht]
    \centering
    \caption{Convolutional backbone configuration ($\sim$1M parameters).}
    \label{tab:conv_config}
    \begin{tabular}{lc}
        \toprule
        Hyperparameter & Value \\
        \midrule
        Model channels $C$ & 128 \\
        Number of blocks $L$ & 2 \\
        Context dimension & 256 \\
        GroupNorm groups & 8 \\
        \bottomrule
    \end{tabular}
\end{table}

\subsection{Training Details}
\label{sec:appendix_training}

\paragraph{Optimizer.}
We use the Muon optimizer \citep{muon2024}, a variant of momentum SGD with Newton-Schulz orthogonalization applied to the momentum buffer. Muon has shown strong performance on transformer training with fewer hyperparameters to tune than Adam variants.

\paragraph{Diffusion Sampling.}
At inference, we use 128 denoising steps with confidence-based token ordering \citep{tokenordering2025}, which unmasks tokens in order of prediction confidence margin.

\subsection{Training Configuration}
\label{sec:appendix_training_config}

\Cref{tab:training_config} summarizes the training hyperparameters used for all experiments. We train for a maximum of 60,000 steps with early stopping based on validation loss, using a patience of 30 validation checks (i.e., training stops if the validation loss does not improve for 30 consecutive checks). The large batch size of 1,100 is enabled by BF16 mixed-precision training, which reduces memory consumption while maintaining numerical stability.

\begin{table}[ht]
    \centering
    \caption{Training hyperparameters.}
    \label{tab:training_config}
    \begin{tabular}{lc}
        \toprule
        Hyperparameter & Value \\
        \midrule
        Optimizer & Muon \\
        Learning rate & $2 \times 10^{-4}$ \\
        Momentum & 0.95 \\
        Max steps & 60,000 \\
        Batch size & 1,100 \\
        Warmup steps & 1,000 \\
        Precision & BF16 mixed \\
        Gradient clipping & 1.0 \\
        Early stopping patience & 30 \\
        \bottomrule
    \end{tabular}
\end{table}

\subsection{Loss Variants}
\label{sec:appendix_loss}

We train three model variants with different loss functions, all derived from the continuous-time MDLM objective. Let $p_\theta(x \mid x_t)$ denote the model's predicted distribution over tokens, $\sigma_t$ the noise level, and $w_t = \frac{d\sigma/dt}{\exp(\sigma_t) - 1}$ the time-dependent weight.

\paragraph{Full Model.}
The standard loss computes cross-entropy over all 10 colors at every position:
\begin{equation}
\mathcal{L}_{\text{full}} = \mathbb{E}_{t, x_t} \left[ -w_t \cdot \log p_\theta(x_0 \mid x_t) \right]
\end{equation}

\paragraph{Occupancy Model.}
The occupancy loss uses binary cross-entropy between background (color 0) and foreground (colors 1--9). We first compute the foreground probability by summing over all non-background colors:
\begin{align}
p_{\text{bg}} &= p_\theta(x = 0 \mid x_t) \\
p_{\text{fg}} &= \sum_{c=1}^{9} p_\theta(x = c \mid x_t)
\end{align}
The binary target is $y = \mathbf{1}[x_0 > 0]$, and the loss is:
\begin{equation}
\mathcal{L}_{\text{occ}} = \mathbb{E}_{t, x_t} \left[ -w_t \cdot \log p_\theta(y \mid x_t) \right]
\end{equation}
where $p_\theta(y \mid x_t) = p_{\text{fg}}$ if $y = 1$ else $p_{\text{bg}}$.

\paragraph{Color Model.}
The color loss computes cross-entropy only at foreground positions, masking out background pixels:
\begin{equation}
\mathcal{L}_{\text{color}} = \mathbb{E}_{t, x_t} \left[ -w_t \cdot \log p_\theta(x_0 \mid x_t) \cdot \mathbf{1}[x_0 > 0] \right]
\end{equation}
This trains the model to predict colors only where objects exist, ignoring the background entirely.

\paragraph{Per-Expert Calibration.}
The occupancy and colour specialists are trained on disjoint label subspaces, so their raw confidence margins are not directly comparable across tokens outside each expert's objective. We therefore allow a fixed per-expert temperature $T_m$ before routing, i.e., logits or log-probabilities are divided by $T_m$ before computing the confidence score in \cref{eq:pixelmax}. We consider this calibration a property of how each expert is prepared, and the routing temperatures used for the reported per-pixel results are:

\begin{table}[ht]
    \centering
    \caption{Per-expert routing temperatures for FactorDiff maximum selection. $T=1.0$ means no calibration; $T<1$ sharpens the expert's confidence margins before routing.}
    \label{tab:routing_temperatures}
    \begin{small}
    \begin{tabular}{lcc}
        \toprule
        Expert pair & Expert 1 temperature & Expert 2 temperature \\
        \midrule
        Full + Full & Full: 1.0 & Full: 1.0 \\
        Occ + Full & Occ: 1.0 & Full: 1.0 \\
        Col + Occ & Col: 1.0 & Occ: 0.25 \\
        \bottomrule
    \end{tabular}
    \end{small}
\end{table}




\subsection{Majority-Voting Baseline}
\label{sec:appendix_compute}

Per-pixel routing executes the $K$ forward passes per denoising step that any composition method already requires (one per expert); the routing layer itself adds only a top-2 sort over the vocabulary and an arg-max over the $K$ confidences, both negligible beside a single transformer forward pass. We therefore take $K$ passes as the floor and report the overhead of alternatives in \cref{tab:compute}. To verify that the improvement comes from routing rather than from extra compute, \cref{tab:selfconsistency} compares per-pixel max against majority-voting (where we resampling a single expert $N{=}3$ times and take the majority-vote output for each pixel) on a 1{,}200-sample validation subset (10 queries $\times$ 120 tasks). Majority-voting plateaus near each expert's deterministic ceiling ($<0.5$ points even at $3\times$ compute), since additional samples only reduce sampling noise within one architecture; per-pixel max gains $+6.3$ points over the strongest single expert at strictly lower total compute, by exploiting the complementary mistakes of two architectures.

\begin{table}[ht]
    \centering
    \caption{Forward passes per denoising step for $K{=}2$ experts. Per-pixel routing adds only negligible bookkeeping ($\epsilon$) above the per-expert floor.}
    \label{tab:compute}
    \begin{small}
    \begin{tabular}{lcc}
        \toprule
        Method ($K{=}2$ experts) & Fwd / step & Rel.\ to floor \\
        \midrule
        Per-pixel max (ours) & $2+\epsilon$ & \textbf{1$\times$ (floor)} \\
        Majority-voting, one expert ($N{=}3$) & 3 & 1.5$\times$ \\
        Cross-expert majority vote ($N{=}3$ each) & 6 & 3$\times$ \\
        Particle correctors (dFKC, $M{=}8$) & 16 & 8$\times$ \\
        \bottomrule
    \end{tabular}
    \end{small}
\end{table}

\begin{table}[ht]
    \centering
    \caption{Per-pixel max vs.\ Majority-voting (Full+Full, 1{,}200 samples). Resampling a single expert cannot close the gap that routing between complementary architectures opens.}
    \label{tab:selfconsistency}
    \begin{small}
    \begin{tabular}{lcc}
        \toprule
        Method & Compute & Exact (\%) \\
        \midrule
        Trans-IR alone & 1$\times$ & 89.2 \\
        Conv-IR alone & 1$\times$ & 85.8 \\
        Trans-IR majority-voting ($N{=}3$) & 3$\times$ & 89.5 \\
        Conv-IR majority-voting ($N{=}3$) & 3$\times$ & 86.0 \\
        Per-pixel max (ours) & 2$\times$ & \textbf{95.5} \\
        \bottomrule
    \end{tabular}
    \end{small}
\end{table}

\subsection{Training Task Samples}
\label{sec:appendix_samples}

\begin{figure}[p]
    \centering
    \includegraphics[page=1,width=0.85\textwidth]{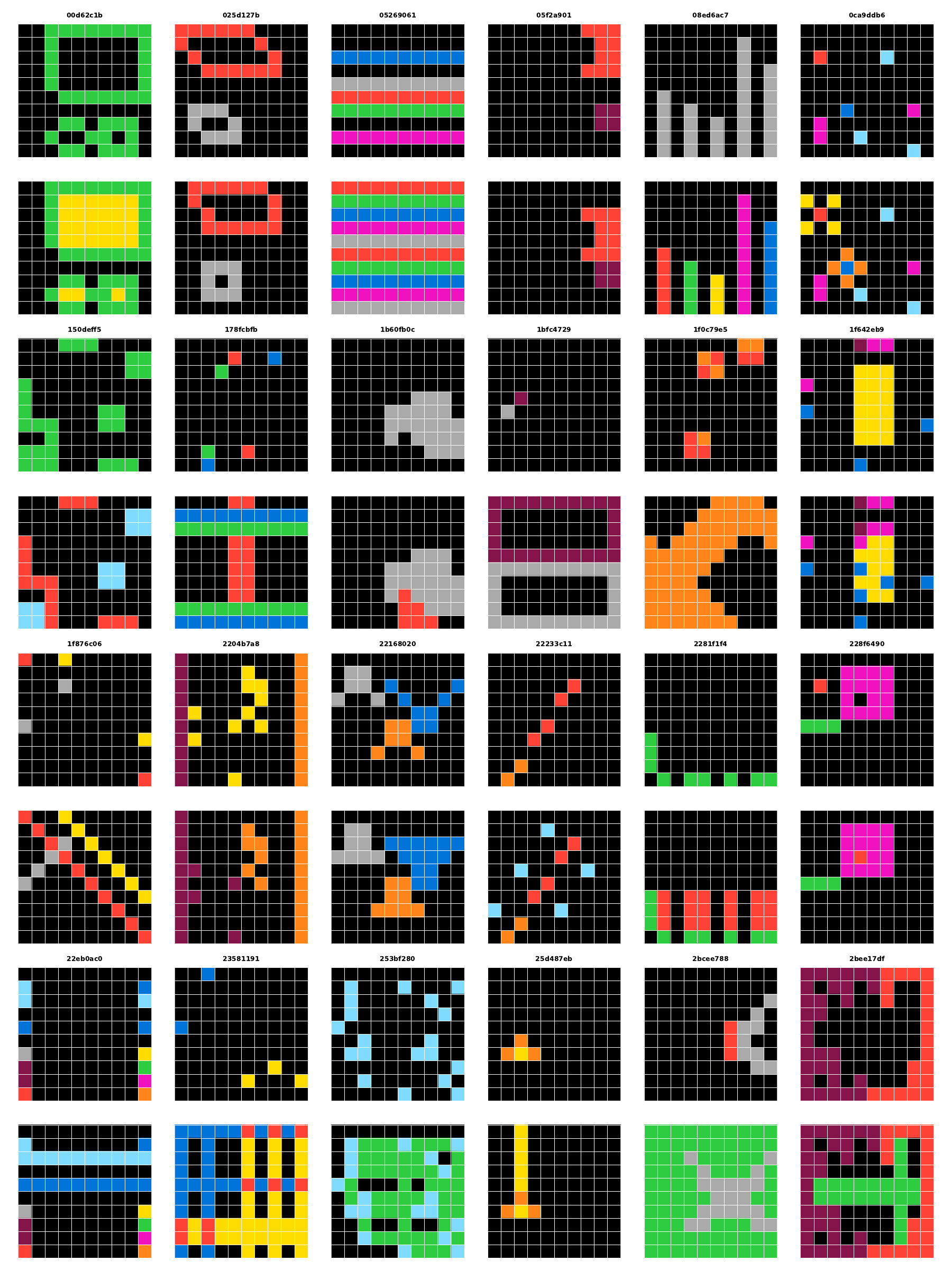}
    \caption{Training task samples (page 1 of 5). Each cell shows an input-output pair from the RE-ARC generator, with task ID labelled above.}
    \label{fig:task_samples}
\end{figure}

We train on 120 tasks from the in-distribution subset of ARC-AGI-1, for which the RE-ARC generator \citep{re_arc2024} provides programmatic augmentation. \Cref{fig:task_samples} shows representative input-output examples produced by these task generators. Each pair displays the input grid (top) and target output grid (bottom), with the task ID labelled above. The diversity of transformations includes pattern completion, object manipulation, symmetry operations, and colour-based rules. All main paper results are evaluated on RE-ARC–generated held-out examples from these same 120 tasks (IID evaluation), not the official ARC-AGI-1 evaluation set.

\begin{figure}[p]
    \centering
    \includegraphics[page=2,width=0.85\textwidth]{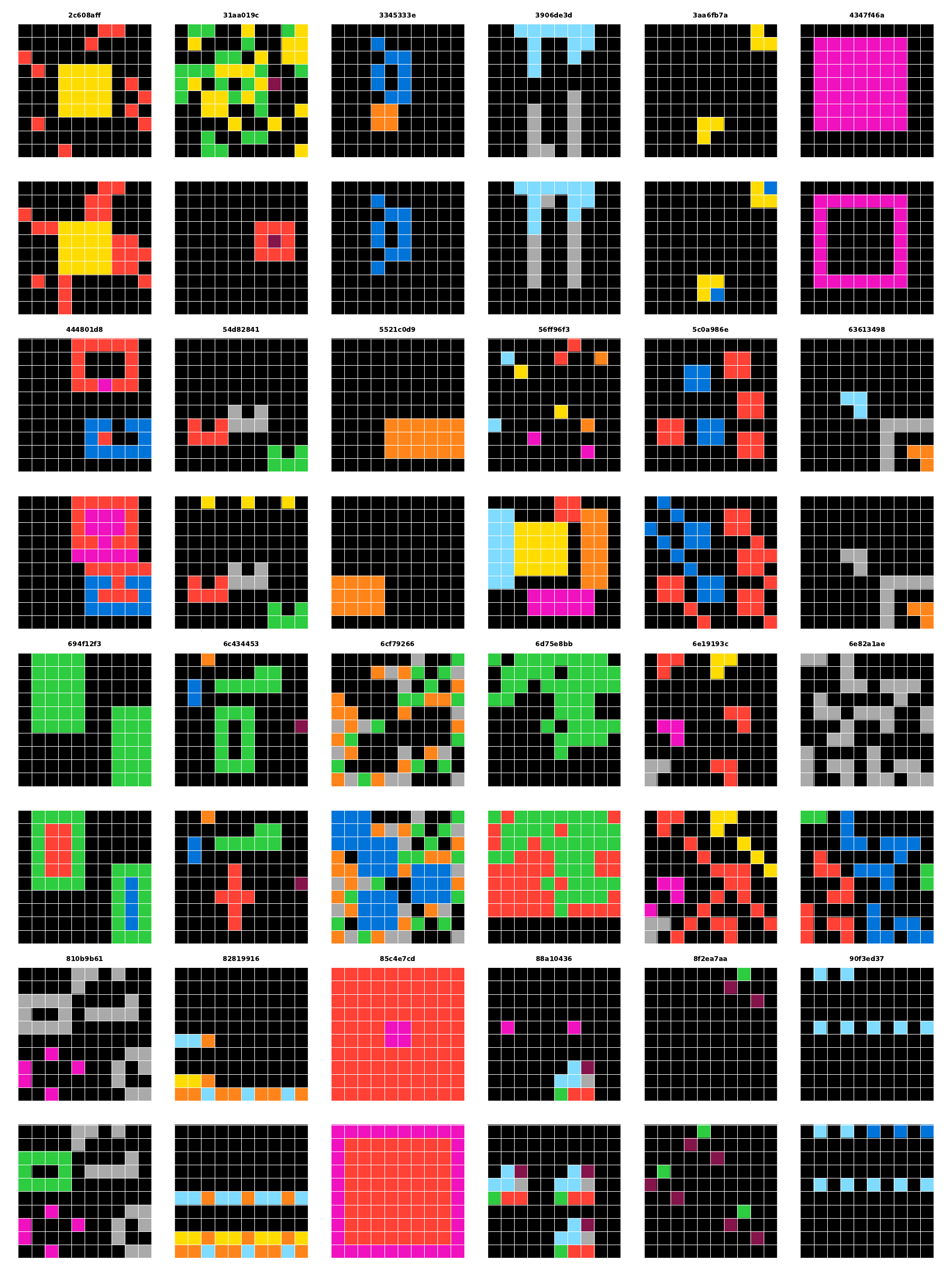}
    \caption{Training task samples (page 2 of 5).}
\end{figure}

\begin{figure}[p]
    \centering
    \includegraphics[page=3,width=0.85\textwidth]{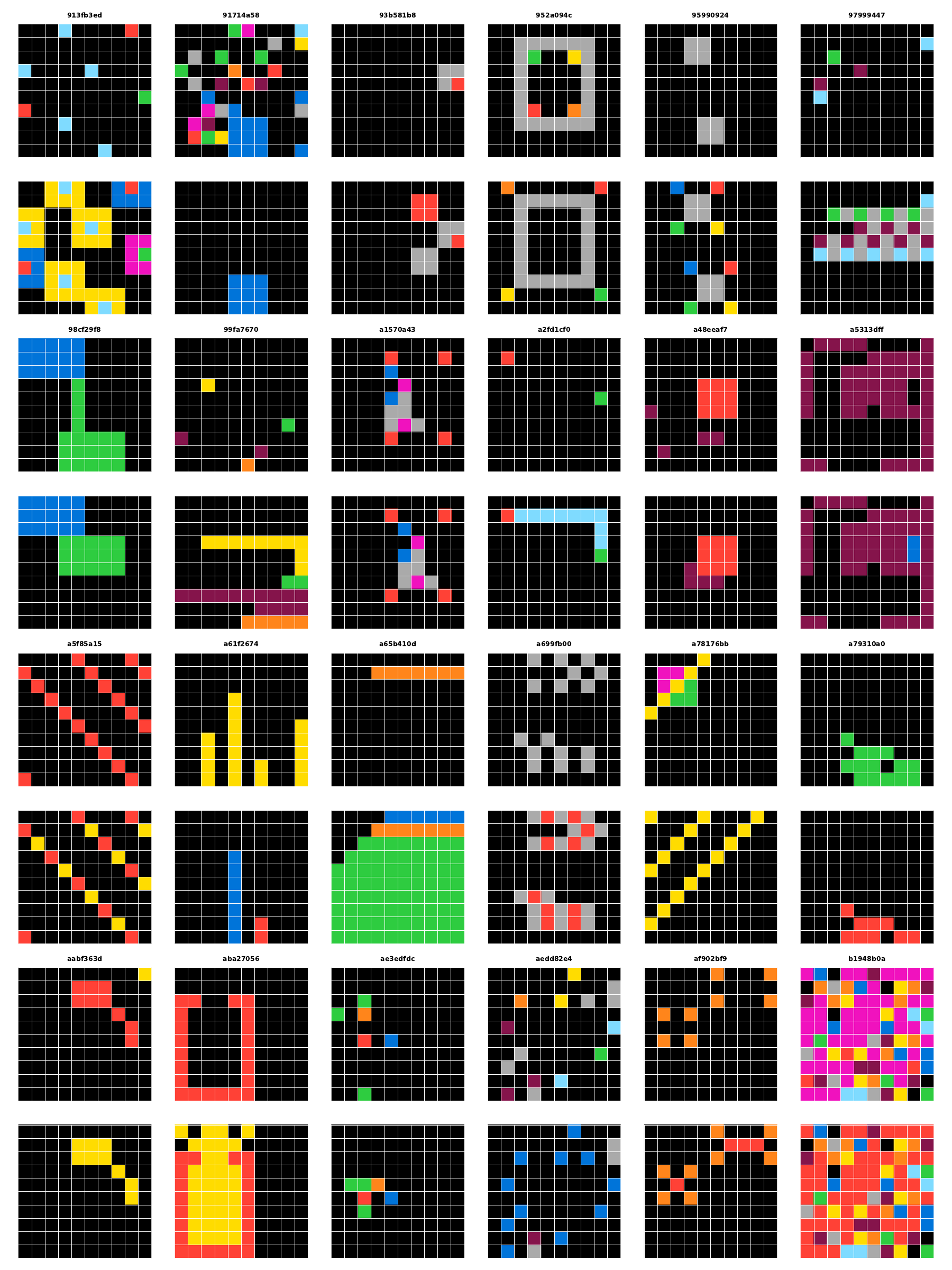}
    \caption{Training task samples (page 3 of 5).}
\end{figure}

\begin{figure}[p]
    \centering
    \includegraphics[page=4,width=0.85\textwidth]{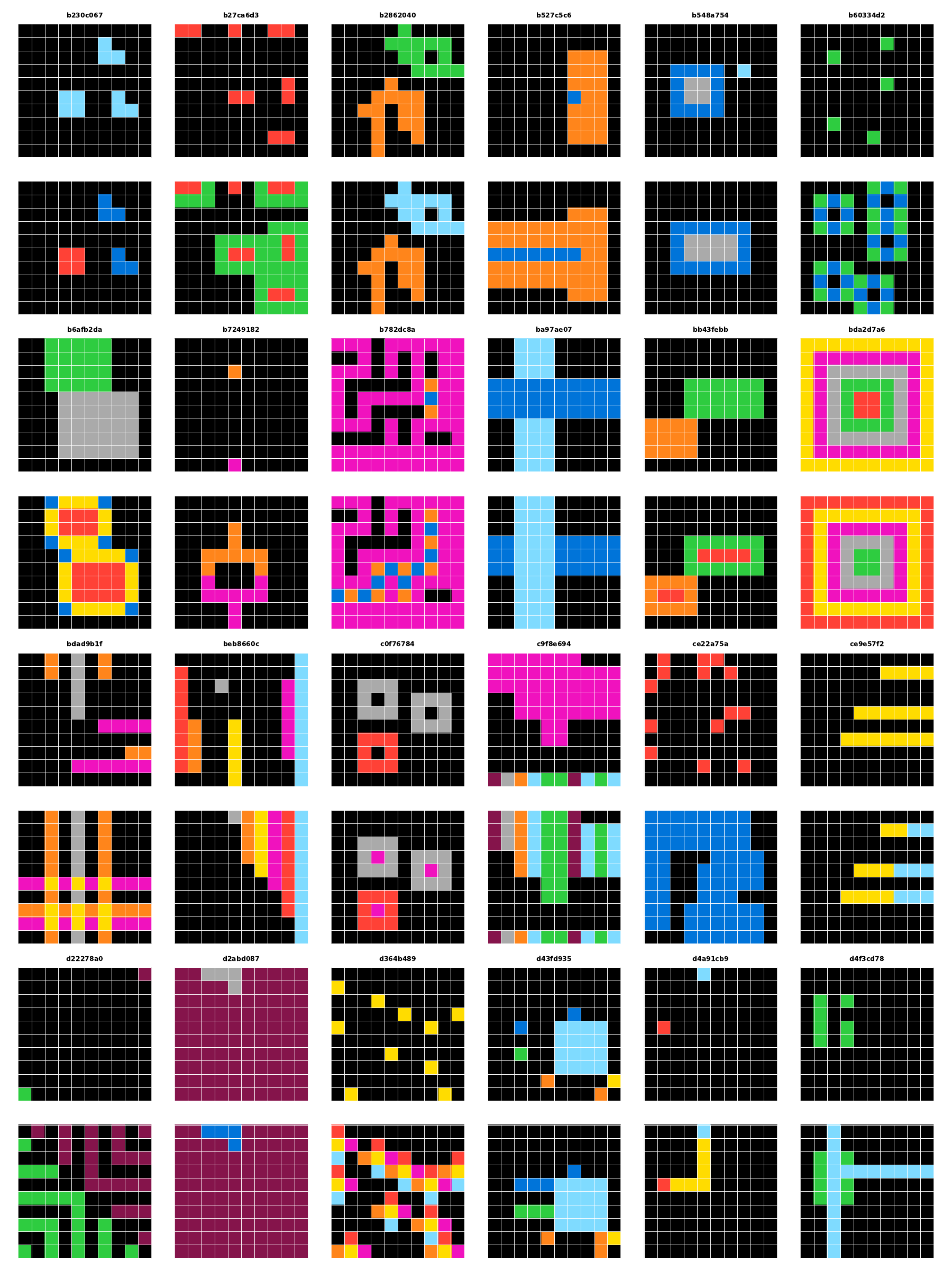}
    \caption{Training task samples (page 4 of 5).}
\end{figure}

\begin{figure}[p]
    \centering
    \includegraphics[page=5,width=0.85\textwidth]{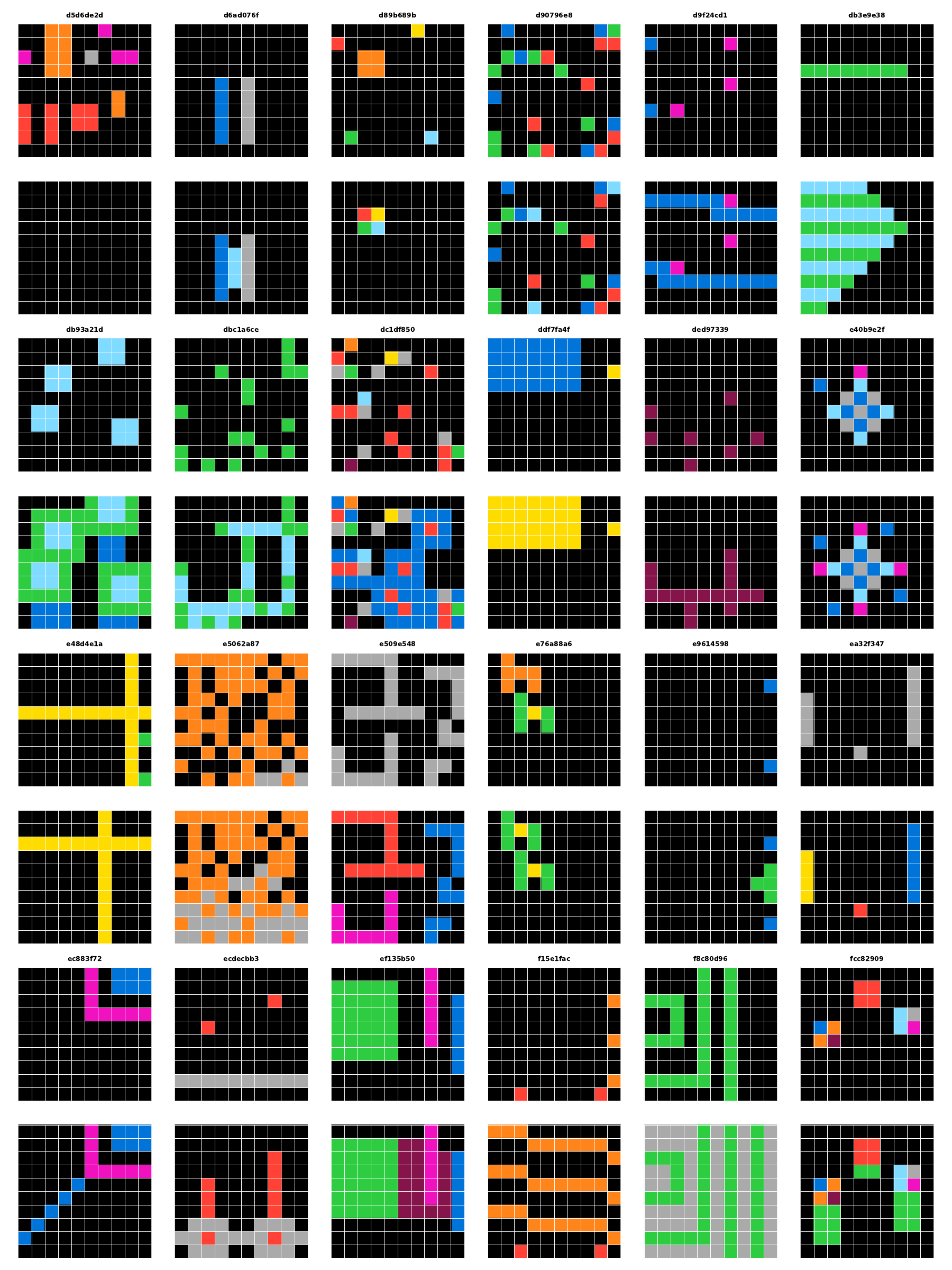}
    \caption{Training task samples (page 5 of 5).}
\end{figure}

\clearpage

\end{document}